\documentclass[conference]{IEEEtran}
\IEEEoverridecommandlockouts
\usepackage{cite}
\usepackage{amsmath,amssymb,amsfonts}
\usepackage{graphicx}
\usepackage{textcomp}
\usepackage{xcolor}
\def\BibTeX{{\rm B\kern-.05em{\sc i\kern-.025em b}\kern-.08em
    T\kern-.1667em\lower.7ex\hbox{E}\kern-.125emX}}

\newcommand{\answerTODO}[1][]{\textcolor{red}{\bf [TODO]}}

\usepackage{multirow}
\usepackage[utf8]{inputenc} 
\usepackage[T1]{fontenc}    
\usepackage{hyperref}       
\usepackage{url}            
\usepackage{nicefrac}       
\usepackage{microtype}      
\usepackage{bm}             
\usepackage{amssymb}
\usepackage{amsmath}
\usepackage{commath}
\usepackage{mathtools}
\usepackage{subcaption}
\usepackage[linesnumbered,ruled]{algorithm2e}
\usepackage{algpseudocode}
\usepackage{array}
\usepackage{booktabs}
\usepackage{tikz}
\usepackage{comment}
\usepackage{amsmath,amssymb} 

\usepackage{url}      
\usepackage{amsfonts}  
\usepackage{nicefrac}     
\usepackage{microtype}  
\usepackage{wrapfig}
\usepackage{bm}
\usepackage{bbm}
\usepackage{mathtools}
\usepackage{listings}
\usepackage{xcolor,colortbl}
\usepackage{comment}

\definecolor{asparagus}{rgb}{0.86, 0.82, 1.0}
\definecolor{darksienna}{rgb}{0.24, 0.08, 0.08}
\definecolor{etonblue}{rgb}{0.59, 0.78, 0.64}
\definecolor{apple}{rgb}{0.66, 0.89, 0.63}

\newtheorem{theorem}{Theorem}

\newtheorem{lemma}{Lemma}

\newcommand{\exv}[2]{\mathbb{E}_{#1}{\left[#2\right]}}

\usepackage[accsupp]{axessibility}

\begin{document}

\title{Robust and Efficient Imbalanced Positive-Unlabeled Learning with Self-supervision}
\author{}
\author{Emilio Dorigatti,$^\dagger$Jonas Schweisthal,$^\dagger$ Bernd Bischl, Mina Rezaei \\
\{emilio.dorigatti,jonas.schweisthal,bernd.bischl,mina.rezaei\}@stat.uni-muenchen.de
\\
Statistical Learning and Data Science, Ludwig-Maximilians-University Munich \\
{\small $^\dagger$ equal contribution}
}

\maketitle

\begin{abstract} ~\label{sec:abstract}
Learning from positive and unlabeled (PU) data is a setting where the learner only has access to positive and unlabeled samples while having no information on negative examples.
Such PU setting is of great importance in various tasks such as medical diagnosis, social network analysis, financial markets analysis, and knowledge base completion, which also tend to be intrinsically imbalanced, i.e., where most examples are actually negatives.
Most existing approaches for PU learning, however, only consider artificially balanced datasets and it is unclear how well they perform in the realistic scenario of imbalanced and long-tail data distribution.
This paper proposes to tackle this challenge via robust and efficient self-supervised pretraining.
However, training conventional self-supervised learning methods when applied with highly imbalanced PU distribution needs better reformulation. 
In this paper, we present \textit{ImPULSeS}, a unified representation learning framework for \underline{Im}balanced \underline{P}ositive \underline{U}nlabeled \underline{L}earning leveraging \underline{Se}lf-\underline{S}upervised debiase pre-training.
ImPULSeS uses a generic combination of large-scale unsupervised learning with debiased contrastive loss and additional reweighted PU loss.
We performed different experiments across multiple datasets to show that ImPULSeS is able to halve the error rate of the previous stat-of-the-art, even compared with previous methods that are given the true prior.
Moreover, our method showed increased robustness to prior misspecification and superior performance even when pretraining was performed on an unrelated dataset.
We anticipate such robustness and efficiency will make it much easier for practitioners to obtain excellent results on other PU datasets of interest.
The source code is available at
\url{https://github.com/JSchweisthal/ImPULSeS}
\end{abstract}


\section{Introduction} \label{sec:intro}

Positive-Unlabeled (PU) learning is a binary classification setting in which only positive and unlabeled samples are given.
Notably, no labeled negative is available, e.g. because of physical limitations of the measurement technology or higher annotation cost.
Learning from such kind of data was first tackled by two-steps~\cite{Chapelle2009} and cost-sensitive approaches~\cite{bekker2020,rezaei2020generativeijcars}.
Recently, unbiased learning paradigms were proposed \cite{du2014analysis,kiryo2017positive,yoo2021accurate} which were later extended by more advanced iterative solutions \cite{chen2020self,dorigatti2022positive}.
These works assume that the positive class prior $\pi=p(y=1)$ is known and treat it as a hyper-parameter, while the most recent approaches incorporate estimation of this class prior in the learning problem in an end-to-end fashion~\cite{chen2020variational,gargMEP}. Debiased learning can potentially introduce new desirable properties to a learned feature space. Nevertheless, unsupervised debiased PU learning has not yet seen such widespread adoption, and it remains a challenging endeavor for representation learning and self-supervision learning paradigms.

Self-supervised contrastive learning is amongst the most promising methods for learning from limited labeled data.
In contrast to supervised approaches, they learn representation without any label annotated labels. 
The main idea of contrastive learning is to contrast semantically similar (positive) and dissimilar (negative) pairs of data points, pulling the representations of similar pairs to be close while simultaneously pushing apart dissimilar pairs. Recent self-supervised contrastive learning algorithms have outperformed even supervised learning~\cite{chen2020big}. However, moving from the controlled benchmark data to uncontrolled real-world data will run into several gaps and challenges. 
For example, most natural image and language data exhibit a long-tail and imbalanced distribution where the frequency of the samples between the different classes is not balanced. 

Indeed, an important problem that has so far been scarcely considered in the academic literature on PU learning, despite being greatly relevant in practice, is such heavily imbalanced setting where the vast majority of unlabeled samples are from the negative class.
Here, similarly to traditional supervised classification approaches when no appropriate measures are taken, most PU learning algorithms struggle to learn to identify positives and in the worst case collapse to constant negative predictions.
A first attempt to circumvent this issue is the imbalanced nnPU loss introduced by \cite{su2021positive}.
This loss is a modified version of the traditional nnPU loss \cite{kiryo2017positive} where the weight of the positive and unlabeled components of the loss are adjusted so that positives are weighted equally as the still unknown negative samples in the unlabeled set.
The accompanying learning-theoretical analysis showed that the excess risk of a classifier trained with such loss decreases as $\mathcal{O}(1/\sqrt{n_p}+1/\sqrt{n_u})$ as is common with other losses for PU data, where $n_p$ and $n_u$ are the number of training positive and unlabeled respectively.
This suggests that, in spite of the reweighting, classifiers trained with low $n_p$ may not reach satisfactory performance even with significantly larger $n_u$.

\begin{figure*}
    \centering
    \includegraphics[width=\linewidth]{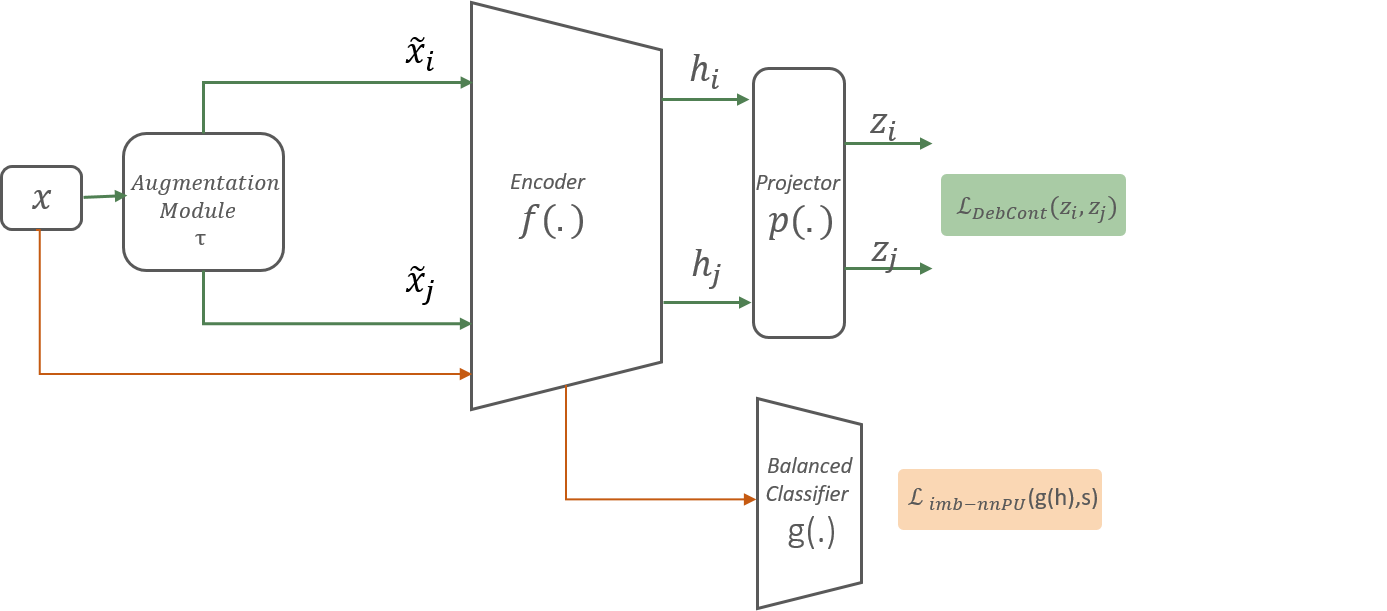} 
   \caption[Training procedure of self-supervised representation learning for imbalanced PU learning]{\small Training procedure of self-supervised representation learning for imbalanced PU learning. \emph{Step 1: Unsupervised pre-training (green)}: From the observations $\bm{x}$, the augmentations $\tilde{\bm{x}}_i$ and $\tilde{\bm{x}}_j$ are generated and processed successively by an encoder $f$ and a projector $p$. Its outputs $\bm{z}_i$ and $\bm{z}_j$ are used to compute a debiased loss with the goal of clustering similar observations. \emph{Step 2: Classifier training (orange)}: After pretraining, the weights of $f$ are frozen and the representations $\bm{h}$ of the original observations $\bm{x}$ are generated. 
   A linear PU classifier $g$ is finally trained on such representations using the imbalanced nnPU loss on the label $s$.}
   \label{fig:impulses_flow} 
\end{figure*}

Hence the inspiration for introducing self-supervised pretraining to imbalanced PU learning: by leveraging the often abundant set of unlabeled examples, it is possible to improve the representations of the positive samples and greatly improve the performance of PU classifiers trained on these representations.
Intuitively, such pretraining shifts the burden of learning good representations from the PU classifier, which has very little information to learn from, to a module that is especially designed to learn in an unsupervised fashion and produce high-quality representations, proven to be beneficial for a variety of downstream tasks.
As we show in our experiments, these representations already capture most of the distinction between positives and negatives in the dataset, thus greatly simplifying the learning task for the PU classifier and making PU learning considerably more robust, efficient and approachable by practitioners in other fields.

To summarise, our contributions are as follows:
\begin{itemize}
\item We are the first to introduce self-supervised pretraining to the imbalanced PU learning setting, and showcase the advantages of our approach in heavily imbalanced scenarios with positive:negative ratios up to 1:11 and labeled:unlabeled ratios up to 1:54.
\item We provide a theoretical analysis showing that the excess risk of our proposed framework has an upper bound that improves with the power of the feature extractor.
\item We show large gains in performance, reducing the error rate by 45\% on the traditional PU benchmark of CIFAR-10 compared to the previous state-of-the-art while using five times fewer labeled positives.
\item We show that pretraining improves PU classification performance over a fully-supervised baseline without pretraining, and is only 0.6 percentage points away from a fully-supervised baseline with pretraining. Moreover it greatly reduces the need to accurately specify the class prior $\pi$ and results in very robust learning even with severe misspecifications.
\end{itemize}

\section{Related works and background} 

\subsection{Positive-unlabeled learning}
PU learning was first proposed in~\cite{Liu2003} and later connected to one-class learning~\cite{ruff2018deep,li2010positive}, multi-positive learning~\cite{xu2017multi,rezaei2017brain,rezaei2018generative}, multi-task learning~\cite{KajiYS18}, and semi-supervised learning~\cite{Chapelle2009}.
Approaches to PU learning can be divided into three categories \cite{bekker2020}: two-step techniques, class prior incorporation, and biased learning, the latter class of approaches being particularly relevant to the present work. A learning-theoretical analysis of PU learning was first proposed by \cite{plessis2014}, who provided an unbiased estimator of the risk under the assumption of known positive class prior probability.
Such estimator was later refined by \cite{Kiryo2017}, who forced a lower bound for the risk to prevent overfitting in a deep learning context.
This approach was recently extended to handle imbalanced datasets \cite{su2021positive}, different kinds of biases in the sampling process \cite{Kato2019,rezaei2020generative,Hsieh2019,Luo2021,apu}, and to remove the need for estimating the positive class prior \cite{chen2020variational,gargMEP}.

\subsection{Self-supervised learning}
Initial works in self-supervised representation learning focused on the problem of learning embeddings without labels such that linear classifier on the learned embeddings could achieve competitive accuracy as supervised model~\cite{doersch2015unsupervised}. Later, self-supervised learning algorithms aim to learn the representation using auxiliary prediction tasks such as image jigsaw puzzles~\cite{noroozi2016unsupervised}, relative patch prediction~\cite{doersch2015unsupervised,doersch2017multi}, image in-painting~\cite{pathak2016context}, image rotation~\cite{komodakis2018unsupervised}, divergence learning~\cite{rezaei2021deep} and tracking moving objects~\cite{wang2015unsupervised}. Self-supervised contrastive learning is amongst the most successful methods to achieving linear classification accuracy and outperforming supervised learning tasks when the network is trained with convolutional-based architectures~\cite{caron2020unsupervised,rezaei2021deep,chen2020simple,chen2020big,chen2020intriguing,zbontar2021barlow}, or vision transformer-based architectures~\cite{wang2021not}. ~\cite{arora2019theoretical} provides theoretically analysis and generalization error bound for the contrastive self-supervised learning framework and pointed sampling bias as one key problem for this learning paradigms. Similarly, Hjelm et al.~\cite{hjelm2018learning} showed that the accuracy of self-supervised contrastive algorithms depends on a large number of negative samples in the training batch.
Recently, sampling bias addressed in pre-training step by combining the contrastive loss and nnpu loss~\cite{} namely \textit{contrastive debiased loss}~\cite{chuang2020debiased,zhao2021graph} or in fine-tuning step \cite{wei2021crest}. 
\cite{yang2020rethinking,liu2021self} pointed out that when the data is imbalanced by class, contrastive learning can learn more balanced feature space than its supervised counterpart. 

In this paper, we study the robustness of representations learned by self-supervised learning and recent debiased contrastive loss on imbalanced PU dataset under a variety of configurations with varying dataset sizes and imbalance ratios.


\section{Method} \label{sec:method}

Our goals are twofold: First, we aim to learn robust representation of imbalanced positive-unlabeled distribution using unsupervised debiased contrastive pre-training. Second, we perform robust semi-supervised classification on top of learned representation via a reweighting of nnpu loss. The training process of our following two-step framework is shown in Figure~\ref{fig:impulses_flow} and described in detail below.

\subsection{Debiased Self-supervised Pretraining}
Given a random image augmentation function $t\sim\mathcal{T}$,
the feature extractor $f_{\bm\theta}$ with parameters $\bm\theta$ takes as input an augmented sample $\tilde{\bm{x}}:=t(\bm{x})$ and produces its latent representations $\bm{h}:=f_{\bm\theta}(\tilde{\bm{x}})$, which is further fed to a projector network $p_{\bm\psi}$ with parameters $\bm\psi$ to create its projection $\bm{z}:=p_{\bm\psi}(\bm{h})$.
Each sample $\bm{x}_i$ in the minibatch $B$ of size $N$ is randomly augmented $M\geq 2$ times (usually $M=2$) and processed by the feature extractor to generate a set of projections $\bm{z}_{ik}:=p_{\bm\psi}(f_{\bm\theta}(t_k(\bm{x}_i)))$ following a random augmentation $t_k\sim\mathcal{T},1\leq k\leq M$.
Such augmented projections are compared among each other via a exponentiated tempered cosine similarity  $s_{ikj\ell}:=\exp(\bm{z}_{ik}^\top\bm{z}_{j\ell})-\exp(\bm{z}_{ik}^\top\bm{z}_{ik})-\exp(\bm{z}_{j\ell}^\top\bm{z}_{j\ell}) - \log\tau$ where $\tau$ is a temperature hyperparameter.
The pretraining procedure aims at increasing the similarity of the augmentations $\bm{z}_{i1},\ldots\bm{z}_{iM}$ among each other and, at the same time, pushing them away from the augmentations $\bm{z}_{jk},j\neq i$.
This is achieved by minimizing the following pairwise debiased contrastive loss:
\begin{equation}
\label{eq:ldeb}
\mathcal{L}_{deb}(\bm{z}_{ik},\bm{z}_{j\ell})
=-\log\frac{
\tau^{-1} \cdot s_{ikj\ell}
}{
\tau^{-1} \cdot s_{ikj\ell}
+  d_u(\bm{z}_{ik})/(M-1)}
\end{equation}
where the term $d_u(\bm{z}_{ik})$ incorporates the distance $s_{ik\cdot\cdot}$ of $\bm{z}_{ik}$ with respect to all other augmentations in the batch.
To prevent overfitting, the the theoretical lower bound $\exp(-\tau^{-1})$ of such distance $\tilde{d}_u(\bm{z}_{ik})$ is explicitly enforced:
\begin{equation}
\label{eq:du}
\begin{aligned}
d_u(\bm{z}_{ik}) 
= & \max \bigg\{ \exp(-\tau^{-1}),
\frac{1}{1-\tau^{+}}
\tilde{d}_u(\bm{z}_{ik})
\bigg\}
\end{aligned}
\end{equation}
where
\begin{equation}
\begin{aligned}
\tilde{d}_u(\bm{z}_{ik})=&
\frac{1}{M(N-1)}
\underset{j\neq i}{\sum_{j=1}^N}\sum_{\ell=1}^M
s_{ikj\ell} 
-\tau^{+} \frac{1}{M-1}
\underset{\ell \neq k}{\sum_{\ell=1}^M}
s_{iki\ell}
\end{aligned}
\end{equation}
Pretraining of $f$ and $p$ is then conducted by minimizing the average debiased loss of all projections generated from the minibatch via stochastic gradient descent:
\begin{equation}
\bm\theta^*,\bm\psi^*=\arg\min_{\bm\theta,\bm\psi}
\frac{1}{MN}
\underset{i\neq j}{\sum_{i=1}^N\sum_{j=1}^N}
\sum_{k=1}^M\sum_{\ell=1}^M
\mathcal{L}_{\text{deb}}(\bm{z}_{ik},\bm{z}_{j\ell})
\end{equation}

\subsection{Debiased Imbalanced PU Semi-supervised Learning}
The PU classifier $g_{\bm\phi}$ parameterized by $\bm\phi$ takes as input the representations generated by the feature extractor $f_{\bm\theta^*}$ and is trained by minimizing the imbalanced nnPU loss \cite{su2021positive}, which is suitable when the number of labeled positives is much smaller than the number of unlabeled samples.
This loss re-weights the positive samples such that their loss constitutes a portion $\pi'$ of the total loss (usually $\pi'=1/2$).
Note that the projection network $p$ is discarded after pretraining as it was empirically observed that it is more effective to learn downstream tasks using shallower representations produced by $f$ \cite{chen2020simple}.
A minibatch $B\subset\mathbb{N}$ contains indices of samples $(\bm{x}_i,s_i)$, where $s_i=1$ if $\bm{x}_i$ is positive and $s_i=0$ if unlabeled.
Using the Sigmoid loss $\ell(y,s):=1/(1+\exp(-ys))$,
we define the loss 
restricted respectively to positive and unlabeled samples
for a given label $s$
as 
$\ell^+_s(B):=1/\sum_{i\in B} s_i \cdot \sum_{i\in B : s_i=1} \ell(g_{\bm\phi}](f_{\theta^*}(\bm{x}_i),s))$ and 
$\ell^-_s(B):=1/\sum_{i\in B} (1-s_i) \cdot \sum_{i\in B : s_i=0} \ell(g_{\bm\phi}(f_{\theta^*}(\bm{x}_i),s))$.
Using these definitions, the classifier $g$ is trained by minimizing the following loss via stochastic gradient descent:
\begin{equation}
\label{eq:imbnnpu}
\mathcal{L}_{imbnnPU}(B)=
\pi'\ell^+_1(B)
+ \frac{1-\pi'}{1-\pi} 
\max 
\bigg\{0,
\ell^-_0(B)-\pi\ell^+_0(B)\bigg\}
\end{equation}
Note that the feature extractor $f$ is fixed to the same parameters $\bm\theta^*$ throughout training of $g$, and no random augmentation is applied in this step.

With our theoretical analysis we connect the excess risk of the linear PU classifier learned in the second step with the debiased loss achieved during pretraining.

Let $\mathcal{F}$ be the class of functions used for the feature extractor $f$, let $\hat{f}$ be the empirical minimizer of the debiased contrastive loss $\mathcal{L}_{deb}$, and let $f^*$ be the optimal feature extractor minimizing the expected loss over all datasets from some fixed data distribution.
Similarly, let $\mathcal{G}$ be the class of functions used for the PU classifier $g_f$ that takes as input the representations of a fixed $f$.
Here we focus on linear binary classifiers returning logit scores for both classes, i.e. $g_f(\bm{x})=\bm{W}f(\bm{x})$, thus let $\hat{g}_f$ and $g^*_f$ be the empirical and global risk minimizer of the balanced PN risk (Eq.~8 and 4 in~\cite{shu2020learning}).
Furthermore, fixing the marginal distributions $p(x|y=1)$ and $p(x|y=-1)$, we consider the family of binary classification tasks $\mathcal{C}$ obtained by uniformly choosing the prior $p(y=1)$.
We now define the supervised risk of a feature extractor $f$ for a task $C\sim\mathcal{C}$ as the minimum risk under the softmax loss in a fully-supervised setting that can be achieved with its representations: $\mathcal{R}_{supervised}(f,C)=\inf_{g_f\in\mathcal{G}}\mathcal{R}_{softmax}(g_f,C)$. and also define its counterpart in expectation over all possible tasks: $\mathcal{R}_{supervised}(f)=\exv{C\sim\mathcal{C}}{\mathcal{R}_{supervised}(f,C)}$.
We then make use of the following two results, which hold for any $f$ each with probability $1-\delta$
\begin{align}
\label{eq:chuangresk}
\mathcal{R}_{supervised}(\hat{f}) &\leq \mathcal{L}_{debiased}(f)+k_1 \\
\label{eq:suresk}
\mathcal{R}_{balancePN}(\hat{g}_f,C) &\leq \mathcal{R}_{balancePN}(g^*_f,C)+k_2(C) 
\end{align}
where $k_1$ and $k_2$ are constants depending on data size and distribution, function class and $\delta$, Eq.~\ref{eq:chuangresk} was shown in Theorem~5 of~\cite{chuang2020debiased} and Eq.~\ref{eq:suresk} was shown in Theorem~2 of~\cite{su2021positive}.
Eq.~\ref{eq:chuangresk} connects the softmax loss of a multi-class classifier trained on the representations produced by $\hat{f}$, in expectation over all possible class priors, with the lowest constrastive loss that can be achieved during pre-training. 
Eq.~\ref{eq:suresk} instead bounds the risk on a balanced PU distribution of the empirical minimizer $\hat{g}_f$ on the same conditional PU distribution $p(x|y)$ but with different marginals $p(y)$ with the minimum risk that can be achieved in the same setting.
Note that such bound holds for any $f$, as the feature extractor only affects the inputs of $g$.

To unify such results, we need to connect the risk of a PU classifier under the sigmoid loss with the risk of a fully-supervised classifier under the softmax loss.
Fortunately, the risk estimator $\mathcal{R}_{balancePN}$ computed as in Eq.8 of~\cite{su2021positive} is already an unbiased estimator of the fully supervised risk under the sigmoid loss that can be computed on PU data alone, as shown in Theorem~1 of~\cite{su2021positive}.
Thus, we only need to connect this risk with the risk under the softmax loss of the same classifier:
\begin{lemma}
\label{lemma1}
\textit{
Given a fixed $d$-dimensional feature extractor $f\in\mathcal{F}$, a binary classification task $c\sim\mathcal{C}$ and a PU classifier $g_f\in\mathcal{G}$ with $g_f(\bm{x})=\bm{W}f(\bm{x})$ and $\bm{W}\in\mathbb{R}^{2\times d}$, we have
\begin{align*}
\mathcal{R}_{balancePN}(g_f,C)&\leq\mathcal{R}_{softmax}(g_f,C)
\end{align*}
where $\mathcal{R}_{balancePN}$ is the risk under the sigmoid loss $\ell(t,y)=(1+\exp(ty))^{-1}$ with true label $t\in\{-1,1\}$ and predictions $y=|1,-1|g_f(\bm{x})$.
}
\end{lemma}
Having established the correspondence between the sigmoid and the softmax risk for any task $C$, we can pass through the latter risk to connect the former to the risk of the optimal feature extractor, in expectation over the classification tasks.
\begin{lemma}
\label{lemma2}
\textit{
Denote with $\tilde{g}_f$ the $g_f$ that minimizes the softmax risk for a fixed $f$ so that $\mathcal{R}_{supervised}(f,C)=\mathcal{R}_{softmax}(\tilde{g}_f,C)$, then for any $\delta>0$ with probability $1-\delta$ we have
\begin{align*}
\exv{c\sim\mathcal{C}}{\mathcal{R}_{balancePN}(\tilde{g}_{\hat{f}},C)}
\leq\mathcal{L}_{debiased}(f^*)+k_1
\end{align*}
}
\end{lemma}
For the last step of relating the risk of the empirical PU classifier with the risk of the feature extractor, note that the minimizers of the sigmoid and the softmax risk coincide as both functions are monotonically decreasing in the predicted probability $p$ and have a global minimum at $p=1$, i.e. we have $\hat{g}_f=\tilde{g}_f$ for every $f$.
Therefore, our main theoretical result is the following:
\begin{theorem}
\label{theorem1}
\textit{
For any $\delta>0$ with probability $(1-\delta)^2$ we have
\begin{align*}
\exv{c\in\mathcal{C}}{\mathcal{R}_{balancePN}(\hat{g}_{\hat{f}},C)}
\leq
\mathcal{L}_{debiased}(f^*)+k_1+k_2
\end{align*}
with $k_1=\mathcal{O}(1/\sqrt{n_p+n_u})$ and $k_2=\mathcal{O}(1/\sqrt{n_p}+1/\sqrt{n_u})$ where $n_p$ and $n_u$ are the number of positive and unlabeled samples in the dataset.
}
\end{theorem}
This result essentially shows that when the function class $\mathcal{F}$ of the feature extractor is sufficiently rich to capture the variability in the dataset and learn meaningful representations, the PU classifier trained on such representations is likely to perform well, too, especially in the large-data limit.
Although a theoretical comparison of empirical risk minimizers belonging to different classes $\mathcal{G}$ and $\mathcal{G}'$ is infeasible, our results show that a linear classifier trained on contrastive representations perform better than a deep classifier trained on the full inputs.
Moreover, note that, even though our bound also scales as $\mathcal{O}(1/\sqrt{n_p})$, mit is still considerably tighter than the bound of~\cite{su2021positive} due to the much smaller Rademacher complexity of linear classifiers hidden in the constant $k_2$.

\section{Experimental Setup} \label{sec:experiments}

We introduce our experimental settings including augmentation techniques, network architecture, optimization, datasets, and tasks. To empirically compare our proposed framework to existing contrastive models and PU learning framework, we follow standard protocols by self-supervised learning and evaluate the learned representation in classification, semi-supervised tasks and transfer learning to different benchmarks as well as real-world clinical dataset considering different imbalance ratios. All of our experiments are done at a GPU-cluster with 8 DGX-A100 40G GPUs.

\subsection{Image Augmentation} \label{ch:m1:aug}
As depicted in Figure.~\ref{fig:impulses_flow}, each input image is transformed using augmentation module $\mathcal{T}$ twice to create two augmented views $\tilde{\bm{x}}_i$, $\tilde{\bm{x}}_j$. We followed the augmentation module used in SimCLR \cite{chen2020simple}, which consists of the following transformations: (a) random cropping, (b) resizing of the crop to the original image size, (c) random flipping, (d) color distortion consisting of color dropping, where the image is turned to grayscale with a selected probability, and color jittering, doing random changes of brightness, contrast, saturation in the images \cite{howard2013some}, (e) Gaussian blurring, and (f) solarization. The first three transformations (a, b, c) are always applied, while the rest are applied randomly, with some probabilities. The probability is different for each views.

\subsection{Deep Representation Network Architecture} \label{ssl:archi}
Our representation architecture includes the ResNet-50 \cite{he2016deep} as an encoder $f$ followed by a projector network $p$. For representation learning in the pretraining step the default linear projection head of ResNet-50 with $2048 \rightarrow 1000$ dimensions is replaced by the 2-hidden-layers network projection head $p$ with $2048 \rightarrow 2048 \rightarrow 128$ dimensions and non-linear ReLU activation function \cite{nair2010rectified} to calculate the projections $\bm{z}$. For the ResNet architecture, we used the same hyperparameters described in \cite{he2016deep}. The pre-training step for 100 epochs takes 240, 360, 90 minutes on CIFAR-10, CIFAR-100, and Glaucoma dataset respectively while fine-tuning step for 100 Epochs just needs around 2 minutes for each dataset.

\subsection{Optimization} \label{ssl:subsec_optim}
We used a batch size of 128, which led to 256 different samples in the batch using the transformation module described in \ref{ch:m1:aug}. The network is trained with Adam \cite{kingma2014adam} optimizer with a learning rate $\gamma=3e^{-4}$ for 500 epochs in pre-training step. The contrastive debiased loss is trained with $\tau^+=0.1$ and $\tau=0.5$ as suggested in \cite{chuang2020debiased} while the contrastive loss is trained with $\tau=0.5$. We set $\pi'=0.5$ and $\pi=p(y=1|s=0)$ by proportion of positives in the unlabeled samples per dataset for the imbalanced nnPU loss (Eq.~\eqref{eq:imbnnpu}). Later we also show performance with unknown $\pi$. For fine-tuning and in the semi-supervised setting, the linear classifier $g$ is trained for 100 epochs as well with a batch size of 256 and Adam optimizer with the same settings.

\subsection{Evaluation}
We followed the same procedure as most other papers in PU learning, such as \cite{kiryo2017positive}, \cite{chen2020self}, \cite{chen2020variational}, and \cite{dorigatti2022positive}, where performance is reported exclusively on the fully labeled test dataset without including performance on the artificially generated unlabeled training samples. Since the datasets are imbalanced, we report the performance in terms of F1 score and AUC.

\subsection{Datasets and Tasks} ~\label{ssl:data}
We use the following datasets in our experiments: CIFAR-10/100~\cite{krizhevsky2009learning}, long-tailed PU CIFAR-10/100~\cite{su2021positive}, and a real-world clinical dataset for glaucoma detection \cite{diaz2019retinal}.
A summary of these datasets is shown in Table~\ref{tbl:dsets} and a detailed description follows.

\begin{table}[ht]
\centering
\begin{tabular}{ l r r r r r }
\toprule
Dataset & Lab. P & Unlab. P & Unlab. N & P:N & L:U \\
\cmidrule{1-1}\cmidrule(l){2-4}\cmidrule(l){5-6}
CIFAR-10  & 600 & 2,400 & 30,000 & 1:10 & 1:54 \\
CIFAR-100 & 1,000 & 4,000  & 45,000 & 1:9 & 1:49 \\
Glaucoma  & 163 & 650 & 1,190 & 2:3 & 1:11 \\
\bottomrule
\end{tabular}
\caption{\small For each training dataset, number of labeled positives, unlabeled positives and unlabeled negatives, as well as ratio between total positives and negatives and between labeled and unlabeled samples.}
\label{tbl:dsets}
\end{table}

\paragraph{CIFAR-10}
The train dataset of CIFAR-10 consists of 50,000 images from 10 different classes with 5,000 images per class, the test dataset consists of 10,000 images, also class balanced.
In previous studies for PU learning \cite{kiryo2017positive}, \cite{chen2020self}, \cite{chen2020variational}, and \cite{dorigatti2022positive}, the 10 classes were divided into the 2 super classes ``vehicles'' (4 classes) and ``animals'' (6 classes), and one of the two super classes was defined as positive.
Of the positive class, a fraction $c$ was considered positively labeled $s=1$ to mimic the label probability $c=p(s=1|y=1)$, and the remainder was considered unlabeled $s=0$.
However, this setting produces an approximately balanced data set between positives and negatives (2:3), whereas we want to investigate the approach to imbalanced data.
Consequently, we define the ``vehicles'' (4 classes) as the positive class and downsample the positives in the train dataset to only 3,000 samples, resulting in a 1:10 positives : negatives ratio.
Like \cite{su2021positive} we set $c$ to $0.2$, so that we have a total of 600 labeled positives and 32,400 unlabeled observations in the train dataset, resulting in a labeled:unlabeled ratio of 1:54.
In the test dataset, we continue to use the nearly balanced distribution of the two classes, so that evaluation via naive performance metrics such as the \textit{accuracy} is still possible.

\paragraph{CIFAR-100}
this dataset consists of a train dataset with 50,000 images and a test dataset with 10,000 images which can be divided into 100 balanced classes. These classes can be grouped into 20 balanced super classes containing 5 classes each. We define the two similar super classes ``vehicles 1'' and ``vehicles 2'' as positive and the remaining 18 super classes as negative. Thus we achieve a positives : negatives ratio of 1:9 and no downsampling has to be done. We set $c$ to $0.2$ again, so in total there are 1,000 labeled positives and 49,000 unlabeled samples in the train dataset. In this case, the imbalanced ratio also exists in the test dataset, so that metrics suitable for an imbalanced scenario must be used for evaluation, such as the F1-score. 

\paragraph{Glaucoma}
Glaucoma is an eye disease that can lead to blindness. In fundus images showing the retina of patients, in addition to arteries and veins, the optic disc is visible. The optic disc can be divided into optic cup, a bright center, and neuro-retinal rim, a slightly darker area around the center. Here, an abnormal size of the optic cup compared to the optic disc is an indication of glaucoma disease, which should be detected \cite{diaz2019retinal}. 
As dataset, we use the labeled observations of the dataset used by \cite{diaz2019retinal}. This merges several glaucoma datasets \cite{zhang2010origa},
\cite{sivaswamy2014drishti}, \cite{medina2016estimating} \cite{kohler2013automatic} into one, since the individual datasets contain relatively few observations. In total there are 2,357 samples, 956 with glaucoma (positive) and 1401 without glaucoma (negative). In the absence of a test dataset, we randomly select 85\% of the samples as the train dataset and 15\% as the test dataset, and again label $c=0.2$ of the positive samples. The final result is 163 labeled positive and 1,840 unlabeled observations in the train dataset.

\section{Results} \label{sec:results}

We follow standard protocol~\cite{goyal2019scaling} and evaluate our learned representations with supervised learning approaches, semi-supervised setting as well as transfer learning to other tasks.
First, we give an overview of the performance of other PU learning methods and how they compare with our framework compares (\ref{sec:competitors}), then we present the performance of our method on more datasets and baselines (\ref{sec:semisupervised}) and finally we illustrate the results of transfer learning to PU datasets (\ref{sec:transferlearning}). 
All reported results are averaged over three runs.

\subsection{Competitors} \label{sec:competitors}
In PU learning, there are only few approaches so far that focus on the application to imbalanced data with few positive samples. 
\cite{su2021positive} showed that their imbalanced nnPU loss outperformed many other methods in this setting, such as \textit{nnPU} \cite{kiryo2017positive}, \textit{self-PU} \cite{chen2020self}, \textit{SMOTE} \cite{chawla2002smote}, or \textit{SSImbalance} \cite{yang2020rethinking}. 

For the usual balanced scenario, most state-of-the-art methods report their results for CIFAR-10 trained on the full train dataset with the approximately balanced split ``vehicles'' : ``animals'' with ratio $2:3$, but they are not explicitly suitable for imbalanced learning.
As comparison methods, we select \textit{VPU} \cite{chen2020variational}, \textit{PAN}\cite{hu2021predictive}, \textit{Self-PU} \cite{chen2020self} and \textit{PUUPL} \cite{dorigatti2022positive} and use the performance reported in the articles for CIFAR-10 on the same class split ``vehicles'' vs. ``animals''.
All of these methods used the entire training set and labeled either 1,000 or 3,000 of the 20,000 positives, whereas we only use 600 labeled and 3,000 total positives.

\begin{table}[ht] 
\centering
\begin{tabular}{ c c c c c}
\toprule
Method & Labeled P. & Train P. & Tot. Train & Accuracy \\
\cmidrule{1-1}\cmidrule(l){2-4}\cmidrule(l){5-5}
VPU~\cite{chen2020variational} & 3,000 & 20,000 & 50,000 & 89.5 \\
PAN~\cite{Hu2021PredictiveAL} & 1,000 & 20,000 & 50,000 & 89.7 \\
Self-PU~\cite{chen2020self} & 3,000 & 20,000 & 50,000 & 90.8 \\
PUUPL~\cite{dorigatti2022positive} & 3,000 & 20,000 & 50,000 & 91.4 \\
\cmidrule{1-1}\cmidrule(l){2-4}\cmidrule(l){5-5}
imbnnPU~\cite{su2021positive} & \textbf{600} & \textbf{3,000} & \textbf{33,000} & 86.5 \\
ImPULSeS & \textbf{600} & \textbf{3,000} & \textbf{33,000} & \textbf{95.3} \\
ImPULSeS & 3,000 & 20,000 & 33,000 & \textbf{97.2} \\
\bottomrule
\end{tabular}
\caption{\label{tbl:competitors} \small Performance of SOTA-competitors trained on balanced PU CIFAR-10. Best performance and fewest resources needed are \textbf{bold}.}
\end{table}

Table \ref{tbl:competitors} compares the accuracy of our method with the competitors' on the common test set. After unsupervised pretraining, we freeze the base encoder $f$ and train a supervised linear classifier on top of it. In our proposed architecture, the linear classifier is a fully connected layer followed by the imbnnPU loss, which is connected on top of $f$ after removing the MLP head. Based on Table \ref{tbl:competitors}, our method clearly surpasses the other baselines, and can improve the accuracy by \textit{3.9} percentage points compared to the previous best method PUUPL, while using 66\% of the train samples, 15\% of the positive samples, and 20\% of the total labeled samples compared to the other methods.
By using the entire CIFAR-10 dataset as the competitors did we instead reduced the error rate by three times.

\subsection{Baselines}~\label{sec:semisupervised}
We compare the performance of our model, including unsupervised contrastive pretraining of representations followed by PU learning of a linear classifier, with an unsupervised clustering baseline based on biased and debiased SimCLR and an imbalanced PU classifier with no pre-training.
We also include performance for the traditional balanced version of CIFAR-10 to ease comparison with contemporary works (Table~\ref{tab:semi_results}.
The measured performance in top-1 accuracy, F1 score and AUC suggests our method improves over the imbalanced nnPU loss and achieves a sizable performance gain of up to 9 percentage points on CIFAR-10 dataset. 


\begin{table}[ht] 
\centering
\begin{tabular}{ l c c c c}
\toprule
   Dataset &  Method & Accuracy & F1 & AUC\\
  \cmidrule{1-2} \cmidrule(l){3-5}
 \multirow{4}{*}{CIFAR-10} 
  &  SimCLR~\cite{chen2020self}                     & 60.05 & 0.25 & 98.67\\ 
  &  Debiased contrastive~\cite{chuang2020debiased} & 60.03 & 0.15 & 98.93\\ 
  &  Oversampled-nnPU~\cite{su2021positive}         & 86.54 & 83.06 & 93.18\\
  &  ImPULSeS & \textbf{95.25} & \textbf{94.01} & \textbf{98.99}\\
  \cmidrule{1-2} \cmidrule(l){3-5}
 \multirow{3}{*}{CIFAR-10 Bal.} 
  &  SimCLR~\cite{chen2020self}                     & 60.00 & 0.14 & 99.31  \\
  &  Debiased contrastive~\cite{chuang2020debiased} & 60.00 & 0.22 & 99.38 \\ 
  &  ImPULSeS & \textbf{97.15} & \textbf{96.41} & \textbf{99.40}\\
  \cmidrule{1-2} \cmidrule(l){3-5}
 \multirow{4}{*}{CIFAR-100} 
  &  SimCLR~\cite{chen2020self}                    & \textit{90.03} & 0.60 & 96.04\\ 
  &  Debiased contrastive~\cite{chuang2020debiased} & \textit{90.00} & 0.00 & 95.74\\ 
  &  Oversampled-nnPU~\cite{su2021positive}         & \textit{86.68} & 44.06 & 82.87\\ 
  &  ImPULSeS & \textit{89.11} & \textbf{62.64} & \textbf{95.89}\\
  \cmidrule{1-2} \cmidrule(l){3-5}
  \multirow{4}{*}{Glaucoma} 
  & SimCLR~\cite{chen2020self}                    & 55.83 & 0.0 & 64.73\\ 
  & Debiased contrastive~\cite{chuang2020debiased} & 55.83 & 0.0 & 58.03\\ 
  & Oversampled-nnPU~\cite{su2021positive}   & \textbf{75.00} & 67.01 & \textbf{77.70}\\
  & ImPULSeS    &  74.22 & \textbf{68.27} & 77.32\\
 \bottomrule
\end{tabular}
 \caption[]{\label{tab:semi_results}\small
 PU classification performance on the imbalanced datasets as well as balanced CIFAR-10.
  The best score for each metric is printed \textbf{bold}.}
\end{table}

\subsection{Transfer Learning}~\label{sec:transferlearning}
We further assess the generalization capacity of the learned representation via transfer learning.
Here, our representations are trained in an unsupervised manner on the training set of the ImageNet ILSVRC-2012 dataset~\cite{imagenet}, and a linear PU classifier is trained second on this representations (Table~\ref{tab:transfer_results}). The accuracy for CIFAR-10 and CIFAR-100 is either unaffected or improved by a few percentage points, while for the Glaucoma dataset performance is severely decreased.
Unsurprisingly, performing transfer learning without fine-tuning representations was not as effective for training PU classifiers on datasets that were very different from the dataset used for pretraining.


\begin{table}[ht]
\centering
\begin{tabular}{ l c c c}
\toprule
   Dataset &  Accuracy & F1 & AUC\\
  \cmidrule{1-1} \cmidrule(l){2-4}
 \multirow{1}{*}{CIFAR-10} 
  & \textbf{95.07} & \textbf{93.82} & \textbf{98.97}\\
  \cmidrule{1-1} \cmidrule(l){2-4}
 \multirow{1}{*}{CIFAR-100} 
  & \textbf{\textit{91.57}} & \textbf{68.39} & \textbf{97.14}\\
    \cmidrule{1-1} \cmidrule(l){2-4}
 \multirow{1}{*}{Glaucoma} 
  & \textbf{\textit{65.28}} & \textbf{53.18} & \textbf{72.46}\\
 \bottomrule
\end{tabular}
 \caption[]{\label{tab:transfer_results} \small  Imbalanced positive and unlabeled classification performance under transfer learning evaluation. The best score for each metric is printed \textbf{bold}.}
\end{table}
\section{Ablation Studies} \label{sec:ablation}

To build intuition around the behavior and the observed performance of the proposed method, we further investigate the following aspects of our approach in multiple ablation studies: the impact of the pretraining fine-tuning losses and the robustness of the learned representation.

\subsection{Analysis of Loss}

\begin{table}[ht]
\centering
\begin{tabular}{c c c c}
\toprule
Pretraining & Fine-tune Loss & Accuracy & F1 \\
\cmidrule(r){1-2}\cmidrule{3-4}
\multirow{2}{*}{Contrastive Loss}
& nnPU & 93.09 & 90.74 \\
& imbnnPU & 94.26 & 92.50 \\
\cmidrule(r){1-2}\cmidrule{3-4}
\multirow{2}{*}{Debiased Contrastive Loss}
& nnPU & 92.75 & 90.20 \\
& imbnnPU & \textbf{95.29} & 94.05 \\
\bottomrule
\end{tabular}
\caption{\small Analysis of pretraining loss (contrastive and debiased contrastive loss) and fine-tuning loss (nnPU and imbalanced nnPU loss) on the imbalanced CIFAR-10 dataset.}
\label{tbl:loss-bcevspu}
\end{table}
As described in Section~\ref{sec:method}, our proposed framework is trained using unsupervised debiased contrastive loss in order to correct the bias introduced by the imbalanced PU distribution. Thus, we want to quantify how much this debiasing correction affects the final performance.
As shown in Table~\ref{tbl:loss-bcevspu}, there was a minor difference of about 1 percentage point in accuracy when pretraining using the debiased contrastive loss compared to the normal biased loss.

Moreover, the performance of a fully supervised classifier trained on the same data but using the true positive and negative labels is an upper bound on the performance of a PU classfier.
To investigate how close our framework comes to this upper bound and whether pre-training can be helpful to reduce the gap between PU and fully supervised performance, we trained both models in a supervised setting on the true labels using a weighted binary cross-entropy (wBCE) loss on the predictions $\hat{y}$:
\begin{equation}
\mathcal{L}_{wBCE}(\hat{y}, y)=w_{pos} y_{i}  \log (\hat{y})
+(1-y_{i}) \log (1-\hat{y})
\end{equation}
where $w_{pos}$ is the weight of the positive samples and is set to the ratio of number of negatives to number of positives in the training dataset.

Table~\ref{tab:self_PUvsSup} shows the result of the pretraining and classification losses.
The debiased$+$wBCE variant achieved better results than training with the wBCE loss from scratch. Notably, the gap between PU learning and fully supervised learning is almost closed by our use of pretraining.
For example, the difference in F1-score and AUC for CIFAR-10 (CIFAR-100) with pretraining was 0.9\% and 0.4\% (4.2\% and 1.5\%), whereas without pretraining the difference was 5.3\% and 3.2\% (9.2\% and 5.3\%).
Importantly this suggests that self-supervised pretraining is not only beneficial to tackle the problem of class imbalance, but also reduces the performance gap between PU and PN data.

\begin{figure}
    \centering
    \begin{subfigure}[t]{0.98\linewidth} 
      \includegraphics[width=\linewidth]{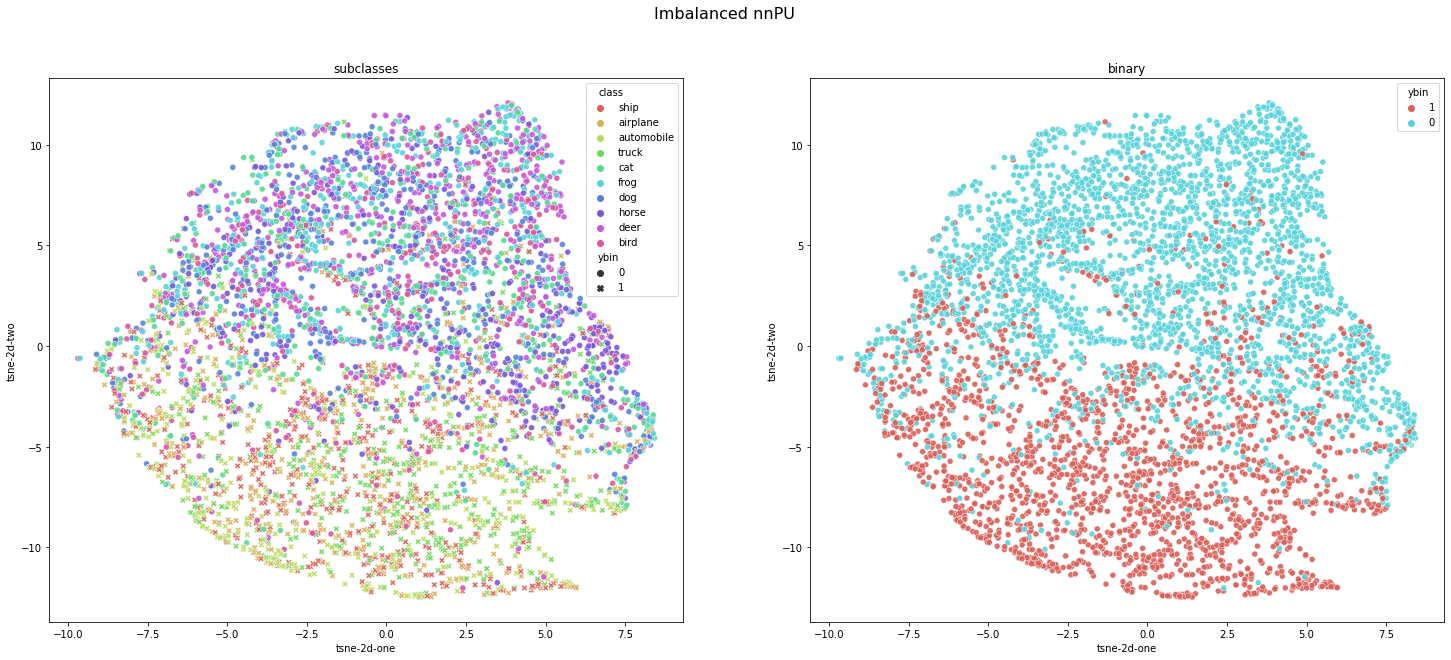} 
   \end{subfigure} 
    \begin{subfigure}[t]{0.98\linewidth} 
      \includegraphics[width=\linewidth]{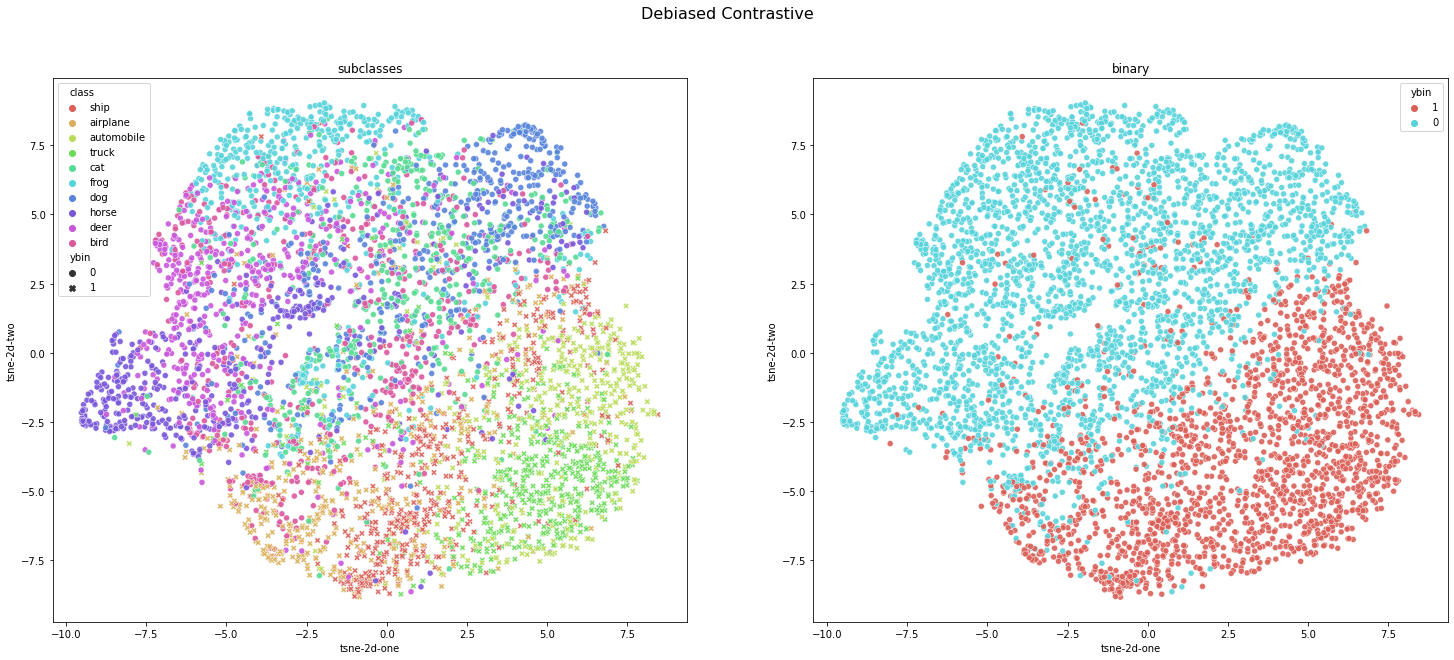} 
   \end{subfigure} 
   \caption[t-SNE visualization of representations of self-supervised pre-training.]{\small t-SNE visualization of representations on test dataset of CIFAR-10. Top: ResNet-50 trained on imbalanced nnPU loss without pre-training. Bottom: After pre-training on debiased contrastive loss. Color-coded for underlying 10 subclasses (left) and binary classes (right).}
   \label{fig:tsne_10_imbnnpu} 
\end{figure}
\begin{table}[ht]
\centering
\begin{tabular}{ c c c c c c}
  \toprule
 Dataset & Labels &  Method & Acc. & F1 & AUC\\
  \cmidrule(r){1-3}\cmidrule{4-6}
 \multirow{4}{*}{CIFAR-10} & \multirow{2}{*}{PU} &   imbnnPU  & 86.5 & 83.0 & 93.6\\
  & & ImPULSeS    & \textbf{95.3} & \textbf{94.0} & \textbf{99.0} \\
  \cmidrule(r){2-3}\cmidrule{4-6}
  &  \multirow{2}{*}{PN} &   wBCE  & 91.0 & 88.3 & 96.8\\
  & & debiased + wBCE   & \textbf{95.9} & \textbf{94.9} & \textbf{99.4}  \\
  \cmidrule(r){1-3}\cmidrule{4-6}
 \multirow{4}{*}{CIFAR-100} & \multirow{2}{*}{PU} &  imbnnPU  & \textit{86.7} & 44.1 & 82.9\\
  & & ImPULSeS    & \textbf{89.1} & \textbf{62.6} & \textbf{95.6} \\
  \cmidrule(r){2-3}\cmidrule{4-6}
  &  \multirow{2}{*}{PN} &   wBCE  & \textit{91.0} & 53.3 & 88.2\\
  & & debiased + wBCE   & \textbf{91.7} & \textbf{68.8} & \textbf{97.1} \\
  \bottomrule
\end{tabular}
 \caption[Performance with and without debiased pre-training for PU vs. supervised data.]{\label{tab:self_PUvsSup}\small Performance with and without debiased pre-training for PU and fully-supervised (PN) labeled data. Best performance in \textbf{bold}.}
\end{table}

\begin{figure}
    \centering
    \includegraphics[width=0.88\linewidth]{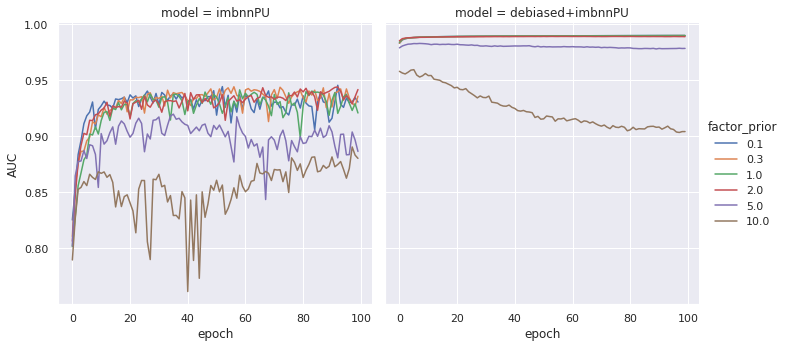} 
   \caption[Robustness against prior mis-specification.]{\small Robustness against prior mis-specification for different distortion factors $b_{dis}$ of the prior. AUC of the test dataset of CIFAR-10 is plotted over 100 training epochs for imbnnPU (left) and for the classifier of debiased+imbnnPU after finished pre-training (right).}
   \label{fig:priormis} 
\end{figure}

\subsection{Quality of Learned Representations}

The main feature of ImPULSeS compared to the competitors is that representation learning is decoupled from the actual classification task. In the second step, ImPULSeS uses the same architecture as the imbnnPU baseline, except that the latter starts training from scratch whereas our method only fine-tunes the weights in the last layer, while keeping the previous layers frozen.
Since our method achieved much better results in the evaluation, the quality of the representation before the last linear layer was obviously significant for the quality of the classifier. 

Figure \ref{fig:tsne_10_imbnnpu} shows the t-SNE \cite{van2008visualizing} visualizations of the 2048-dimensional representations of the test dataset of CIFAR-10 of both models.
Even though the debiased pretraining was completely unsupervised, it resulted in representations that more clearly separated positives and negatives compared to training a classifier from scratch using the imbalanced nnPU loss. Such representations even separated the different classes found in the original dataset, unlike those resulting from the imbalanced nnPU loss that were oblivious to such latent structures. Explicitly representing these substructures makes it then considerably easier to identify the unlabeled positive samples even when very little labeled examples are provided.
It is for this reason that decoupling representation learning from classifier learning was found to indeed improve the final classification performance \cite{kang2019decoupling}, and the robustness of contrastive self-supervised pretraining on imbalanced data has been well studied \cite{yang2020rethinking}, \cite{liu2021self}.
This principle is also found in few-shot learning \cite{wang2020generalizing}, where a classifier is also trained using representation learning and few labeled data.
Our contribution is to show that this robustness is also highly beneficial for PU learning.

Such high quality representations already separating classes in the latent space raise the question of whether a PU loss is needed at all.
We thus trained a classifier using the BCE loss and treating all unlabeled as negative samples.
While the equally-weighted BCE failed to learn any signal and collapsed to all-negative predictions due to the data imbalance, the weighted BCE performed better than the nnPU loss and was very competitive with the imbalanced nnPU loss.

\subsection{Robustness Against Mis-specification of the Class Prior}

In the previous analyses, we always assumed the class prior $\pi$ to be known, and in the imbnnPU loss for each dataset to be set to the proportion of positive samples in the unlabeled samples, following \cite{su2021positive} and \cite{kiryo2017positive}.
In real-world applications, however, this proportion is often not known and must be estimated using domain knowledge or other methods, and imperfect estimation can degrade the final classification performance.
In the following, we thus investigate how misspecification of $\pi$ in the imbalanced nnPU loss affects the performance of the model by defining distortion factor $b_{dis}$ of the true prior $\pi$, so that in each case the prior $\pi$ was replaced by the distorted prior $\pi_{dis}=b_{dis}\cdot \pi$.
Here, $b_{dis}$ varies from 0.1 to 10.0, with 1.0 yielding the model with the correct $\pi$. 


In Figure \ref{fig:priormis}, the course of the AUC on the test dataset of CIFAR-10 is displayed over the 100 training epochs of the PU classifier with distorted priors for both imbnnPU and ImPULSeS .
In general, our method had higher AUC than the model without self-supervised pretraining even under different misspecified priors.
The variance was also smaller and the learning curve more stable, since only the weights of the classifier and not the entire ResNet-50 parameters were trained.
For both models, no deterioration of the performance was observed when underestimating $\pi$, likely because the portion of positives was already very low.
When overestimating $\pi$, a deterioration could be observed for both models starting at $b_{dis}\geq5$, although the performance decrease of our method was less marked.
At extreme overestimation levels of $b_{dis}=10$, strong deviations in stability occurred for both methods and the AUC of our model even decreased in the course of training.
At the same time, the initial AUC was above 95\% owing to the high-quality representations learned during pretraining, which do not need the parameter $\pi$.
In conclusion, even with strong misspecification of the class prior, high stability and performance were achieved by our framework.

\section{Conclusion}
In this paper, we unified contrastive self-supervised pretraining with PU learning on imbalanced datasets with very few labeled positive samples.

Through different experiments, we showed that such procedure considerably improves classification performance, halving the error rates of previous state-of-the-art and almost closing the gap between PU and fully-supervised learning.
We then showed that our method is insensitive to severe misspecification of the true class prior $\pi$, making complicated and brittle end-to-end procedures that try to estimate such hyper-parameter unnecessary.

We attribute such improvements to the high-quality latent representations learned during pretraining, which are able to capture elaborate latent structures found in the dataset and greatly benefit the subsequent PU learning step.
On the other hand, without such pre-training traditional PU learning methods struggle to extract any meaningful information from the large set of unlabeled samples, and the insufficient number of positives only result in high-variance estimators.

We anticipate that the high performance of our method combined with its simplicity and robustness will enable practitioners in other disciplines to make full use of their PU datasets and generate novel insights in their respective fields of study.


\subsection{Broader Impact and Limitation}
\label{sec:impact-limitations}
The broader impacts of our work may be seen in many real-world applications including (but are not limited to) information retrieval, program debugging, anomaly detection, and financial market analysis.
However, several limitations present themselves. For example, while the use of sample augmentations is very common in the vision domain, it is unclear how to design suitable augmentations for other, arbitrary data modalities.
Another limitation of our model compared to other learning methods (such as supervised learning) is that self-supervised learning can demand more computing resources and training time.
Considering the fact that our proposed method does not require manual annotation -- which is usually very expensive -- we would argue that this trade-off is acceptable. Furthermore, due to limited evaluation of the method on other domains, the benefit of our proposed method in other applications and datasets such as robotics and information retrieval is yet to be investigated.

\subsection*{Acknowledgments}
ED is supported by the Helmholtz Association under the joint research school "Munich School for Data Science - MUDS" (Award Number HIDSS-0006).
M. R. and B. B. were supported by the Bavarian Ministry of Economic Affairs, Regional Development and Energy through the Center for Analytics – Data – Applications (ADA-Center) within the framework of BAYERN DIGITAL II (20-3410-2-9-8).
M. R. and B. B. were supported by the German Federal Ministry of Education and Research (BMBF) Munich Center for Machine Learning (MCML).

\bibliographystyle{IEEEtran}
\bibliography{ref}

\newpage
\onecolumn
\appendix
\label{appendix}
\subsection{Proof of Lemma~\ref{lemma1}}
First, note that a binary classifier with softmax activation can be transformed into one with sigmoid activation giving the same output probability.
Denote by $u$ and $v$ the rows of $\bm{W}$ and with $'$ the dot product, then:
\begin{align}
\sigma((u-v)'x)
=\frac{1}{1+e^{-(u-v)'x}}
&=\frac{e^{u'x}}{e^{u'x}+e^{-u'x+v'x+u'x}}
=\frac{e^{u'x}}{e^{u'x}+e^{v'x}}
=Softmax(u'x, v'x)
\end{align}
thus let $g_\sigma$ and $g_\phi$ denote the single-output sigmoid classifier and double-output softmax classifier, both returning the probability of the positive class.
Second, note that for the same probability $0<p<1$, the sigmoid loss is never larger than the softmax loss:
\begin{align}
\sigma(-p)=\frac{1}{1+e^p}\leq -\log p
\Leftrightarrow
\exp
\underbrace{
\left(
\frac{1}{1+e^p}
\right)
}_{<1}
\leq \underbrace{\frac{1}{p}}_{>1}
\end{align}
where the right part is easily seen to be true.
This extends trivially to the respective risks. In particular, $\mathcal{R}_{balancePN}$ was defined in Eq.~5 of~\cite{su2021positive}, 
and using the fact that $\exv{x}{f(x)}\leq\exv{x}{g(x)}$ if $f(x)\leq g(x)$ for every $x$ we have:
\begin{align}
\mathcal{R}_{balancePN}(g_\sigma,C)
&=p(y=1)\exv{p(x|y=1)}{\sigma(-g_\sigma(x))}
+p(y=0)\exv{p(x|y=0)}{\sigma(-(1-g_\sigma(x)))} \\
&\leq p(y=1)\exv{p(x|y=1)}{-\log g_\phi(x)}
+p(y=0)\exv{p(x|y=0)}{-\log(1-g_\phi(x))} \\
&=\mathcal{R}_{softmax}(g_\phi,C)
\end{align}
where we take $\pi'=1/2$ and $\pi=p(y=1)=C$.
Note that in practice we train the PU classifier using the definition in Eq.~8 of~\cite{su2021positive} which makes use of unlabeled data instead of negatives, however their Theorem~1 shows the two definitions to be equivalent, thus allowing us to bridge PU and PN learning in our derivations.

\subsection{Proof of Lemma~\ref{lemma2}}
Making use of Lemma~\ref{lemma1} and Lemma~4 of~\cite{chuang2020debiased},
we have with probability $1-\delta$
\begin{align}
\exv{c\sim\mathcal{C}}{\mathcal{R}_{balancePN}(\tilde{g}_{\hat{f}},C)}
&\leq
\exv{c\sim\mathcal{C}}{\mathcal{R}_{softmax}(\tilde{g}_{\hat{f}},C)}
\leq
\mathcal{L}_{debiased}(f^*)+k_1
\end{align}
where we slightly abuse notation and use $g$ to indicate either $g_\sigma$ or $g_\phi$ depending on the context.

\subsection{Proof of Theorem~\ref{theorem1}}
Here we use Theorem~2 of~\cite{su2021positive}, which provided a bound for a fixed task $C$.
This can be converted to a result in expectation over $C$ using the same expectation inequality used above:
\begin{align}
\exv{C\sim\mathcal{C}}{
\mathcal{R}_{balancePN}(\hat{g}_f,C)
-\mathcal{R}_{balancePN}(g^*_f,C)
}\leq\exv{C\sim\mathcal{C}}{k_2(C)}
\end{align}
with probability $1-\delta$.
Moreover, note that $\hat{g}_f=\tilde{g}_f$, where the former is the minimizer of the PU risk (and thus the fully-supervised PN risk, as per Theorem~1 of~\cite{su2021positive}) and the latter is the minimizer of the softmax risk for a given task $C$.
Since both losses are decreasing in the predicted logits, have a global minimum at $p=1$ and there is a monotonic transformation between the two, the respective minimizers correspond.
Thus, using the above with Lemma~\ref{lemma2} we have
\begin{align*}
\exv{c\in\mathcal{C}}{\mathcal{R}_{balancePN}(\hat{g}_{\hat{f}},C)}
&=
\exv{c\in\mathcal{C}}{\mathcal{R}_{balancePN}(\tilde{g}_{\hat{f}},C)}
\leq
\exv{c\in\mathcal{C}}{\mathcal{R}_{balancePN}(g^*_{\hat{f}},C)} +\exv{C\sim\mathcal{C}}{k_2(C)} \\
&=\exv{c\in\mathcal{C}}{\mathcal{R}_{balancePN}(\tilde{g}_{\hat{f}},C)} +\exv{C\sim\mathcal{C}}{k_2(C)} \\
&\leq\mathcal{L}_{debiased}(f^*)+k_1+k_2
\end{align*}
with probability $(1-\delta)^2$ since both bounds hold with probability $1-\delta$ and we assume they are independent of each other.

\end{document}